\documentclass{article}

\usepackage{PRIMEarxiv}

\usepackage{hyperref}
\usepackage{url}
\usepackage{enumitem}
\usepackage{multirow}
\usepackage{graphicx}
\usepackage{wrapfig}
\usepackage{tabularx}
\usepackage{adjustbox}
\usepackage{tabularx}  
\usepackage{booktabs}  
\usepackage{caption}  
\usepackage{makecell}
\usepackage{color}
\usepackage[authoryear, round]{natbib}

\usepackage[utf8]{inputenc} 
\usepackage[T1]{fontenc}    
\usepackage{hyperref}       
\usepackage{url}            
\usepackage{booktabs}       
\usepackage{amsfonts}       
\usepackage{nicefrac}       
\usepackage{microtype}      
\usepackage{lipsum}
\usepackage{fancyhdr}       
\usepackage{graphicx}       
\graphicspath{{media/}}     
\usepackage{amsmath,bm}

\usepackage{wrapfig}

\title{Image Super-Resolution via Latent Diffusion: A Sampling-Space Mixture of Experts and Frequency-Augmented Decoder Approach
}

\author{
  Feng Luo \thanks{Equal Contribution.}  \\
  Tencent AI Lab \\
  Shenzhen, China \\
  \texttt{amandaaluo@tencent.com}
   \And
  Jinxi Xiang $^{*}$   \\
  Tencent AI Lab \\
  Shenzhen, China \\
  \texttt{jinxixiang@tencent.com}
   \And
  Jun Zhang \thanks{Corresponding author.} \\
  Tencent AI Lab \\
  Shenzhen, China \\
  \texttt{junejzhang@tencent.com}
   \And
  Xiao Han \\
  Tencent AI Lab \\
  Shenzhen, China \\
  \texttt{haroldhan@tencent.com}
   \And
  Wei Yang \\
  Tencent AI Lab \\
  Shenzhen, China \\
  \texttt{willyang@tencent.com}
}

\begin{document}
\maketitle

\begin{abstract}
The recent use of diffusion prior, enhanced by pre-trained text-image models, has markedly elevated the performance of image super-resolution (SR). To alleviate the huge computational cost required by pixel-based diffusion SR, latent-based methods utilize a feature encoder to transform the image and then implement the SR image generation in a compact latent space. Nevertheless, there are two major issues that limit the performance of latent-based diffusion. First, the compression of latent space usually causes reconstruction distortion. Second, huge computational cost constrains the parameter scale of the diffusion model. To counteract these issues, we first propose a frequency compensation module that enhances the frequency components from latent space to pixel space. The reconstruction distortion (especially for high-frequency information) can be significantly decreased. Then, we propose to use Sample-Space Mixture of Experts (SS-MoE) to achieve more powerful latent-based SR, which steadily improves the capacity of the model without a significant increase in inference costs. These carefully crafted designs contribute to performance improvements in largely explored 4$\times$ blind super-resolution benchmarks and extend to large magnification factors, i.e.,  8$\times$ image SR benchmarks. The code is available at \href{https://github.com/tencent-ailab/Frequency_Aug_VAE_MoESR}{https://github.com/tencent-ailab/Frequency\_Aug\_VAE\_MoESR}.
\end{abstract}


\section{Introduction}

Diffusion models have quickly emerged as a powerful class of generative models, pushing the boundary of text-to-image generation, image editing, text-to-video generation, and more visual tasks\citep{SohlDickstein2015DeepUL, Song2019GenerativeMB, Song2020ScoreBasedGM, Ho2020DenoisingDP}. In this paper, we explore the potential of diffusion models to tackle the long-standing and challenging image super-resolution (SR) task.

Let us revisit the diffusion model in the context of generation space. Early diffusion models, operating in the high-dimensional pixel space of RGB images, demand substantial computational resources. To mitigate this, the Latent Diffusion Model (LDM)\citep{Rombach2021HighResolutionIS} uses VQGAN to shift the diffusion process to a lower-dimensional latent space, maintaining generation quality while reducing training and sampling costs. Stable Diffusion further enhances LDM\citep{Rombach2021HighResolutionIS} by scaling up the model and data, creating a potent text-to-image generator that has garnered significant attention in the generative AI field since its release. 

However, a significant challenge arises when dealing with higher compression rates, drastically affecting detail consistency. As noted in studies \citep{Kim2020ZoomtoInpaintII, Rahaman2018OnTS}, the convolutional nature of autoencoders tends to favor learning low-frequency features due to spectral bias. So the escalation in compression rate leads to loss of visual signals in the high-frequency spectrum, which embodies the details in pixel space. While some image synthesis researches \citep{Lin2023CatchMD, Zhu2023DesigningAB} have recognized and addressed these issues, they have received little attention in the field of super-resolution\citep{Wang2022ZeroShotIR, Chung2022DiffusionPS, Lin2023DiffBIRTB}. StableSR\citep{Wang2023ExploitingDP} is one of the few models that fine-tune the autoencoder decoder with the CFW module, offering a potential solution to this problem in the spatial domain.

Regarding the training of the diffusion-based SR model, one approach\citep{wang2022zero, chung2022diffusion, kawar2022denoising} involves using the pre-trained Stable Diffusion model, incorporating certain constraints to ensure fidelity and authenticity. However, the design of these constraints presupposes knowledge of the image degradations, which are typically unknown and complex. As a result, these methods often demonstrate limited generalizability. Another approach to address the above challenge involves training a super-resolution (SR) model from scratch, as seen in studies \citep{Saharia2021ImageSV, Li2021SRDiffSI, Rombach2021HighResolutionIS, Sahak2023DenoisingDP}. To maintain fidelity, these methods use the low-resolution (LR) image as an additional input to limit the output space. Although these approaches have achieved significant success, they often require substantial computational resources to train the diffusion model, especially when the dimension exceeds $512\times 512$ or $1024\times 1024$. In this context, the super-resolution models are relatively small, possessing fewer parameters than the image generation model.  

To this end, we aim to enhance the diffusion model for SR by fixing the decoding distortion and enlarging the diffusion model capacity without significantly increasing computational cost. We first propose a frequency compensation module that enhances the frequency components from latent to pixel space. The reconstruction distortion can be significantly decreased by better aligning the frequency spectrums of the high-resolution and reconstructed images. Then, we propose to use the Sample-Space Mixture of Experts (SS-MoE) to achieve stronger latent-based SR, which steadily improves the capacity of the model in an efficient way. Our approach allows for enlarging the model size without incurring significant computational costs during training and inference.

In summary, we highlight our contributions in three aspects:
\begin{itemize}[leftmargin=0.04\linewidth]
    \item We identify the issue of information loss within the latent diffusion model used for image SR. In response, we propose a frequency-compensated decoder complemented by a refinement network. This innovative approach is designed to infuse more high-frequency details into the reconstructed images, thereby enhancing the overall image quality.
    
    \item We design sampling-space MoE to enlarge the diffusion model for image SR. This allows for enhanced high-resolution image processing without necessitating a substantial increase in training and inference resources, resulting in optimized efficiency. 

    \item We evaluate the model on $4\times$ Blind SR and $8\times$ Non-Blind SR benchmarks, employing both quantitative and qualitative assessment methods. Additionally, we conduct essential ablation studies to validate the design choices of the models. Experiment results show that we achieved solid improvement in terms of perceptual quality, especially in $8\times$ SR.
\end{itemize}

\section{Related Work}

\textbf{Image SR.}
 Image SR aims to restore an HR image from its degraded LR observation. Recent advancements \citep{liu2022blind} in Blind Super-Resolution (BSR) have delved into more intricate degradation models to mimic real-world deterioration. Specifically, BSRGAN \citep{Zhang2021DesigningAP} is designed to emulate more realistic degradations using a random shuffling approach, while Real-ESRGAN \citep{Wang2021RealESRGANTR} leverages "high-order" degradation modeling. Both methodologies employ Generative Adversarial Networks (GANs) \citep{Goodfellow2014GenerativeAN, Miyato2018SpectralNF}  to understand the image reconstruction process amidst complex degradations. FeMaSR \citep{Chen2022RealWorldBS} interprets  SR as a feature-matching issue, utilizing the pre-trained VQ-GAN \citep{Esser2020TamingTF}. Despite the utility of BSR techniques in mitigating real-world degradations, they fall short of generating realistic details.

\textbf{Diffusion Model for Image SR.}
Using diffusion models in image SR signifies a burgeoning trend. The primary driving force behind this methodology is the exceptional generative capacity of diffusion models. Numerous research efforts have explored their use in image restoration tasks, specifically enhancing texture recovery \citep{Ramesh2022HierarchicalTI, Rombach2021HighResolutionIS, Nichol2021GLIDETP}. The remarkable generative prowess of these pre-trained diffusion models has been showcased, underscoring the critical need for high-fidelity inherent in SR.
Based on the training strategy, these studies can be broadly classified into two categories: supervised training and zero-shot methods. The first category \citep{Saharia2021ImageSV, Li2021SRDiffSI, Niu2023CDPMSRCD, Sahak2023DenoisingDP} is committed to optimizing the diffusion model for SR from the ground up through supervised learning. The zero-shot approach \citep{Choi2021ILVRCM, Wang2022ZeroShotIR, Chung2022DiffusionPS, Fei2023GenerativeDP} aims to leverage the generative priors in the pre-trained diffusion models for SR, by imposing certain constraints to ensure image fidelity. Those zeros-shot approaches usually show limited ability to super-resolve and the supervised methods are constrained to limited model scale due to huge computation costs.

\textbf{Autoencoder in Diffusion Model.}
To reduce the training and sampling costs linked to the diffusion model, StableDiffusion \citep{Rombach2021HighResolutionIS} spearheads the use of DM-based generation in the latent space. This is achieved specifically through a pre-training phase for an autoencoder model \citep{Esser2020TamingTF}, defined by an encoder-decoder architecture, to navigate the perceptual space proficiently. However, the high compression rate in the latent space often results in image distortion during the reconstruction of images from the low-dimensional latent space \citep{Lin2023CatchMD, Zhu2023DesigningAB}. The lossy latent necessitates a more robust decoder to offset the information loss.

\section{Methodology}

\subsection{Preliminaries}

Given a dataset of low-resolution and target image pairs, denoted as $\mathcal{D}=\left\{\boldsymbol{x}_i, \boldsymbol{y}_i\right\}_{i=1}^N$ drawn from an unknown distribution $p(\boldsymbol{x}, \boldsymbol{y})$. Image SR is a process of conditional distribution modeling $p(\boldsymbol{y} | \boldsymbol{x})$, and it is a one-to-many mapping in which many target images may be consistent with a single low-resolution image.  Our objective is to learn a parametric approximation to $p(\boldsymbol{y} | \boldsymbol{x})$ through a stochastic iterative refinement process that transforms a source image $\boldsymbol{x}$ into a target image $\boldsymbol{y}$.  We tackle this problem by adapting the diffusion probabilistic (DDPM) model\citep{Ho2020DenoisingDP, SohlDickstein2015DeepUL}  to conditional image SR.

DDPM is the first diffusion-based method introduced in \citep{SohlDickstein2015DeepUL}, which consists of a diffusion process and a denoising process. In the diffusion process, it gradually adds random noises to the data $x$ via a T-step Markov chain \citep{Kong2021OnFS}. The noised latent variable at step t can be expressed as:
\begin{equation}
    \mathbf{z}_t=\sqrt{\hat{\alpha}_t} \boldsymbol{y}+\sqrt{1-\hat{\alpha}_t} \epsilon_{\mathbf{t}}, \text{with}\ \hat{\alpha}_t=\prod_{k=1}^t \alpha_k \quad \epsilon_{\mathbf{t}} \sim \mathcal{N}(\mathbf{0}, \mathbf{1}),
    \label{eq:diffusion_forward}
\end{equation}
where $\alpha_t\in(0,1)$ is the corresponding coefficient. For a $T$ that is large enough, e.g., $T=1000$, we have $\sqrt{\hat{\alpha}_{T}}\approx0$ and $\sqrt{1-\hat{\alpha}}_{T}\approx1$. And $\mathbf{z}_{T}$ approximates a random Gaussian noise. Then, the generation of $\mathbf{x}$ can be modeled as iterative denoising. 

\citep{Ho2020DenoisingDP} connect DDPM with denoising score matching and propose a $\epsilon-$prediction form for the denoising process:
\begin{equation}
    \mathcal{L}_t=\left\|\epsilon_t-f_\theta\left(\mathbf{z}_t, \boldsymbol{x}, t\right)\right\|^2,
    \label{eq:diffusion_eps}
\end{equation}
where $f_{\theta}$ is a denoising neural network parameterized by $\theta$, and $\mathcal{L}_t$ is the training loss function. The cornerstone of this design is the denoising neural network, which is typically a UNet. 

 \begin{figure}
    \centering
    \includegraphics[width=0.8\linewidth]{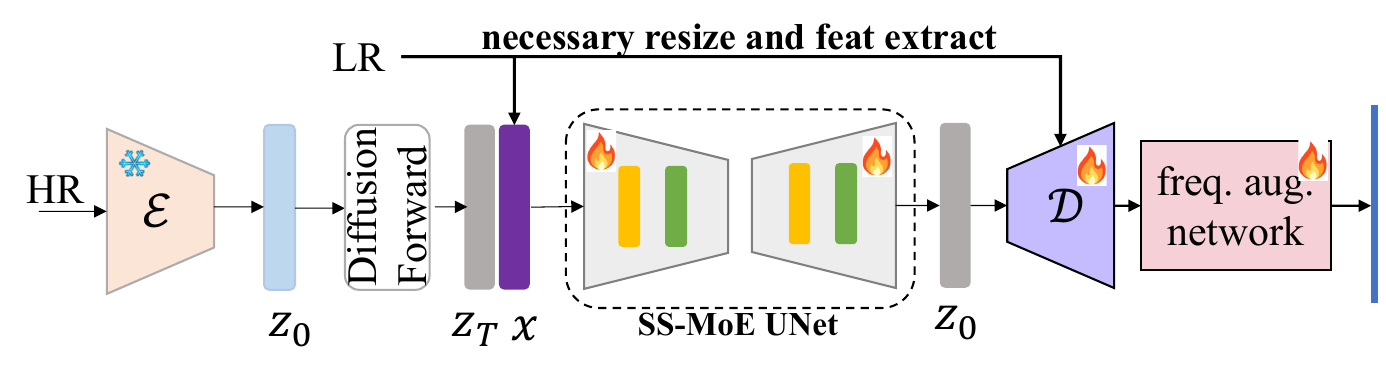}
    \caption{Latent diffusion model for image SR with SS-MoE and frequency augmented decoder.}
    \label{fig:framework}
\end{figure}

During inference, we reverse the diffusion process through iterative refinement, taking the form of:
\begin{equation}
    \boldsymbol{y}_{t-1} \leftarrow \frac{1}{\sqrt{\alpha_t}}\left(\boldsymbol{y}_t-\frac{1-\alpha_t}{\sqrt{1-\gamma_t}} f_\theta\left(\boldsymbol{x}, \boldsymbol{y}_t, \gamma_t\right)\right)+\sqrt{1-\alpha_t} \boldsymbol{\epsilon}_t, \quad \boldsymbol{y}_T\sim \mathcal{N}(\mathbf{0}, \boldsymbol{I}).
    \label{eq:reverse}
\end{equation}

The proposed latent diffusion model for image SR is illustrated in Fig. \ref{fig:framework}. It consists of multiple SS-MoE UNet and a frequency compensated autoencoder that will be illustrated in \ref{sec_3.2} and \ref{sec_3.3}.

\subsection{Denoising UNet with SS-MOEs}
\label{sec_3.2}
\begin{figure}
    \centering
    \includegraphics[width=0.8\linewidth]{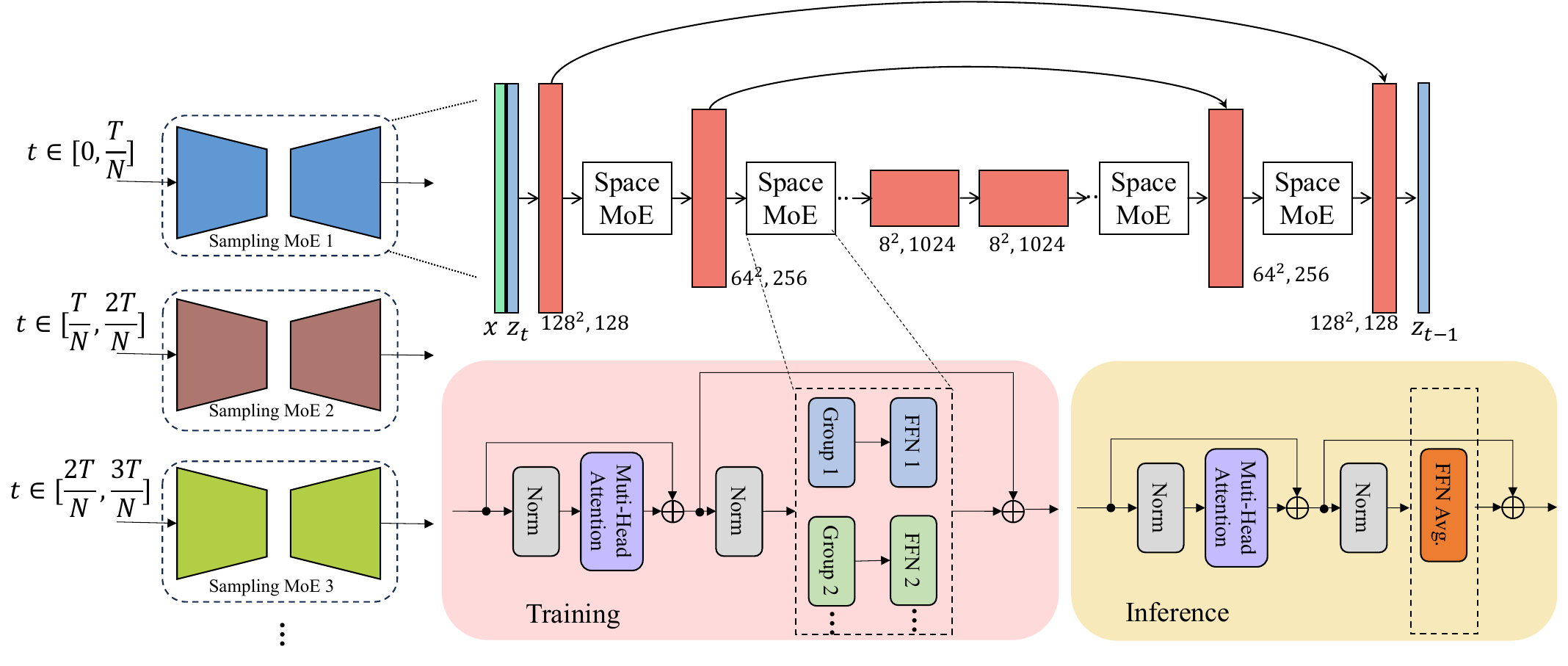}
    \caption{Sampling-Space MoE of the denoising UNet for image SR. "FFN Avg." means averaging the weights of all experts into one FFN for inference after training.}
    \label{fig:ss-moe}
\end{figure}

The denoising UNet structure, denoted as $f_\theta$, is inspired by the Latent Diffusion Model (LDM) \citep{Rombach2021HighResolutionIS}, and it incorporates residual and self-attention blocks as its core building elements. To make the model conditional on the input $\boldsymbol{x}$, we employ bicubic interpolation \citep{Saharia2021ImageSV} to up-sample the low-resolution image to match the target resolution. The up-sampled result is concatenated with $\boldsymbol{z}_t$ along the channel dimension, see Fig. \ref{fig:ss-moe}.

\textbf{Sampling MoE.}
The quality of the image can be significantly improved by utilizing a time-mixture-of-experts (time-MoE) method, a concept derived from earlier studies \citep{Xue2023RAPHAELTG, Feng2022ERNIEViLG2I, Balaji2022eDiffITD}. Similarly, diffusion-based SR is 
also a diffusion process that progressively introduces Gaussian noise to an image over a sequence of timesteps, $t = 1, . . . , T$. The image generator is trained to reverse this process given an upsampled low-resolution image as the condition, denoising the images from $t = T$ to $t = 1$. Each timestep is designed to denoise a noisy image, gradually converting it into a clear high-resolution image. It is important to note that the complexity of these denoising steps fluctuates based on the level of noise present in the image. For instance, when $t = T$, the input image $\boldsymbol{x}_t$ for the denoising network is heavily noisy. However, when $t = 1$, the image $\boldsymbol{x}_t$ is much closer to the original, aka less noisy image. So we divide all timesteps uniformly into $N$ stages consisting of consecutive timesteps and assign single Sampling Expert to one stage. Since only a single expert network is activated at each step, the scale and capacity of our model 
 can expand with computational overhead remaining the same during inference, regardless of an increase in the number of experts. We use $N = 4$ to assure that all experts can be loaded on a GPU when inference.

\textbf{Space MoE.} 
Using FFN MoE for better performance is widely acknowledged in large language models \citep{lepikhin2020gshard, fedus2022switch, roller2021hash, zuo2021taming} and image classification\citep{riquelme2021scaling, huang2023experts}. Specifically, EWA\citep{huang2023experts} designs MoE in a stuctural re-parameterization way to enhance 2D Classification and 3D Vision, validating its effectiveness in discriminative models. Similarly, we create MoE layers with $N$ spatial experts (i.e., $N$ FFNs) $\{E_1, E_{2},...E_{N}  \}$ after existing multi-head attention to scale the denoising UNet in diffusion-based SR. For a batch of input tokens $(B, L, d)$, where $L=hw$, $B$ denotes the batch size and $h$, $w$, $d$ denote the height, width and channel number of a feature map respectively. Assuming $L$ is divisible by $N$, we randomly split the tokens into $N$ groups and then processed with experts:
\begin{equation}
    \{x_1, x_2, ..., x_L \} \xrightarrow{\text{group split}}  \{X_1, X_2,...,X_N \},\ y=E_i(x).
\end{equation}
Given the weights if $N$ experts $\{W_1, W_2,...,W_N  \}$, weight sharing is performed among all experts during training:
\begin{equation}
    \overline{W_i}= \gamma W_i+(1-\gamma) \overline{W}_{j}\  \text{with}\ \overline{W}_{j} =  \sum_{j \neq i}^N \frac{1}{N-1} W_j,
    \label{eq:ewa}
\end{equation}
where $\overline{W_i}$ denotes the updated weight of the $i$-th expert. 
Conceptually, we update the weight of each expert by averaging the weights of the other experts. The momentum coefficient $\gamma \in [0,1)$ regulates the degree of information exchange among the experts. The momentum update, as shown in Eq. \ref{eq:ewa}, ensures a smoother evolution of each expert. Each expert carries a substantial dropout (i.e., $\frac{N-1}{N}$) and they collectively evolve through momentum updates. A relatively large momentum (e.g., $\gamma = 0.999$) works better than a smaller value (e.g., $\gamma = 0.9$), suggesting that smaller $\gamma$ could probably lead to weight collapse (identical weights across all experts). 

After training, each space MoE layer is converted into an FFN layer by simply averaging the experts:
$
    \text{FFN}=\frac{1}{N} \Sigma_{i=1}^N E_i.
    \label{eq:infer_ewa}
$
In this way, introducing space MoE to augment the denoising UNet only incurs computation overhead of single FFN.

\subsection{Frequency Compensated Decoder}
\label{sec_3.3}

 \begin{figure}
    \centering
    \includegraphics[width=0.8\linewidth]{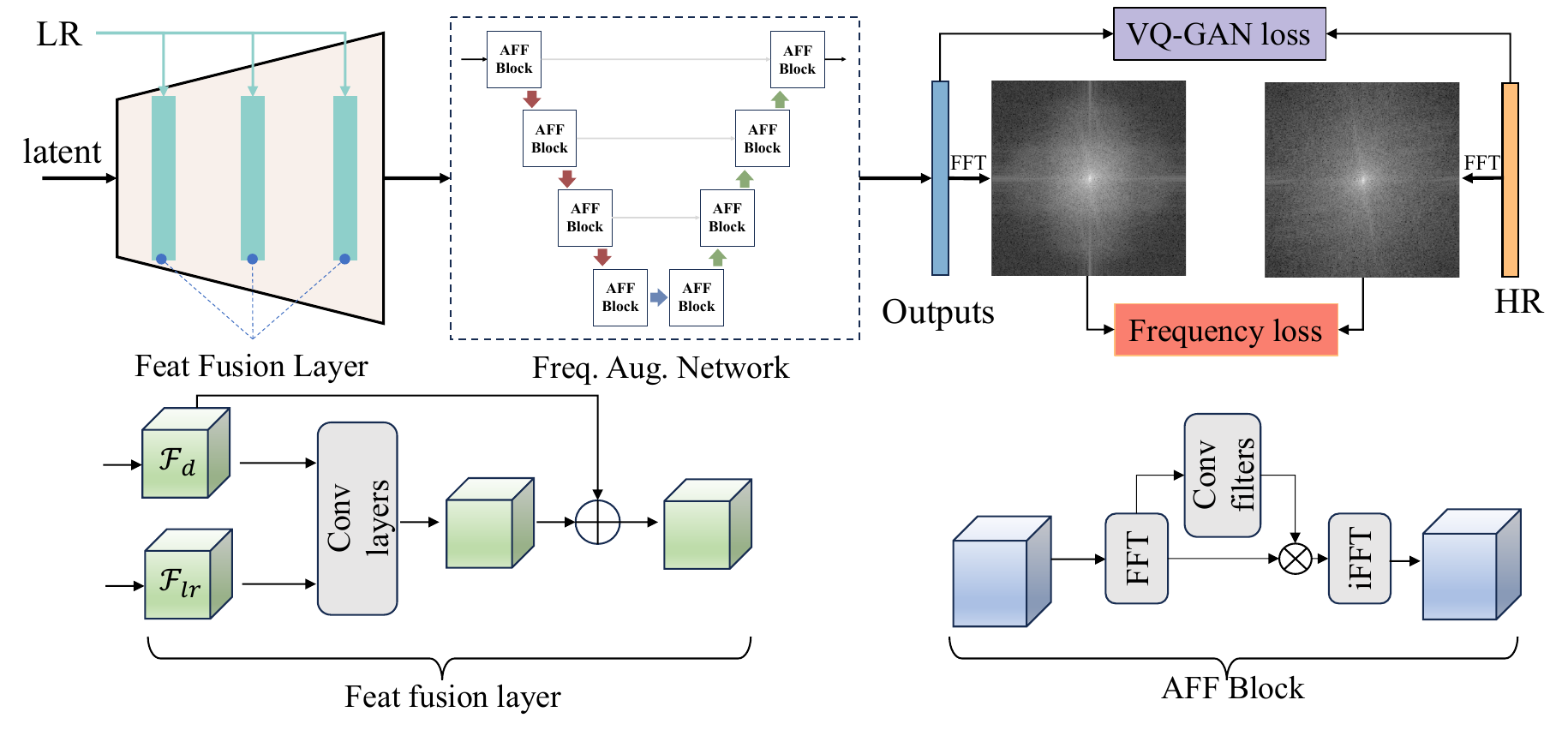}
    \caption{The proposed frequency-augmented decoder is conditioned on the low-resolution image and uses the AFF block \citep{Huang2023AdaptiveFF} to reduce information loss in the frequency domain.}
    \label{fig:decoder}
\end{figure}

To address the information loss using autoencoder, we propose improving the image super-resolution quality by augmenting the decoder with a frequency-compensated loss and network. To be specific, the frequency-augmented decoder comprises a VQGAN\citep{Esser2020TamingTF} decoder conditioning on low-resolution inputs similar to \citep{Wang2023ExploitingDP} and a refinement network utilizing frequency operators along with a frequency loss for optimization, as depicted in Fig. \ref{fig:decoder}.

\textbf{LR-conditioned Decoder.} Conditioning on low-resolution inputs has been proven to enhance the reconstruction fidelity for image SR \citep{Wang2023ExploitingDP, Wang2018RecoveringRT}. We add a feature extractor to get the LR image representation for conditions during decoding. Since only several encoding features are needed, some layers like VQGAN's middle blocks can be dropped to save memory and computation cost during inference.
The fusion of LR features $F_{lr}$ and decode latent $F_{d}$ can be formulated as $F_{m} = F_d + \mathcal{C}{}(F_{lr}, F_d; \theta)$, where $\mathcal{C}{}(\cdot; \theta)$ is a sequences of trainable convolution layers as designed in \citep{Wang2023ExploitingDP}.

\textbf{Refinement Network.}  We further use a tiny UNet model with frequency augmentation operation to address information loss. It is inserted after the last upsample block of the VQGAN decoder. The UNet model consists of six AFF blocks proposed by \citep{Huang2023AdaptiveFF}, ie. sequences of activations, linear layers, and adaptive frequency filters. The frequency operator first transforms the input latent into the frequency domain using the Fourier transform and then applies semantic-adaptive frequency filtering through element-wise multiplication. 

To optimize the frequency-augmented decoder, we use frequency loss $\mathcal{L}_{\text{freq}}$ \citep{Jiang2020FocalFL} in addition to VQ-GAN loss $\mathcal{L}_{\text{VQ-GAN}}$ \citep{Esser2020TamingTF}for reconstruction:
\begin{equation}
    \mathcal{L} = \mathcal{L}_{\text{VQ-GAN}} + \lambda\mathcal{L}_{\text{freq}}\ \text{, with, }   \mathcal{L}_{\text{freq}}=\frac{1}{M N} \sum_{u=0}^{M-1} \sum_{v=0}^{N-1} w(u, v)\left|F_r(u, v)-F_f(u, v)\right|^2.
    \label{eq:ffl}
\end{equation}
We set $\lambda=10$ by default; the matrix element $w(u,v)$ is the weight for the spatial frequency at $(u,v)$; $F_r(u, v), F_f(u, v)$ are the FFT results of ground-truth and reconstruction images.

\section{Experiments}

\textbf{Datasets.}
We train and test our method on 4$\times$ and  8$\times$ super-resolution with synthesized and real-world degradation settings. For each task, there are two training stages, stage 1 for Sampling-Space MoE and stage 2 for Frequency Compensated Decoder. In the first stage, degraded pipelines are different for each task. For the 4$\times$ super-resolution, following StableSR \citep{Wang2023ExploitingDP}, we combine images in DIV2K \citep{agustsson2017ntire}, Flickr2K\citep{timofte2017ntire} and OutdoorSceneTraining\citep{wang2018recovering} datasets as the training set. We additionally add the openImage dataset \citep{kuznetsova2020open} for general cases. LR-HR pairs on DIV2K are synthesized with the degradation pipeline of Real-ESRGAN \citep{Wang2021RealESRGANTR}. The sizes of LR and HR patch are 128$\times$128 and 512$\times$512. For the 8$\times$ super-resolution, we only use DIV2K, Flickr2K and openImage dataset for training. LR images are with a size of 64$\times$64 and obtained via default setting (bicubic interpolation) of Matlab function imresize with scale factor 8. In stage 2 of Frequency Compensated Decoder training, we adopt Sampling-Space MoE to generate $100k$ LR-Latent pairs for 4$\times$ and 8$\times$ SR given the LR images as conditions.

\textbf{Training.}
We train all of our Sampling-Space MoEs for $100k$ steps with a batch size of 144. Moreover, training steps for Frequency Compensated Decoder is $50k$ and the batch size is 32. Following LDM\citep{Esser2020TamingTF}, we use Adam optimizer, and the learning rate is fixed to $5\times 10^{-5}$ and $1\times 10^{-4}$ for SS-MOEs and FCD. All trainings are conducted on 8 NVIDIA Tesla 32G-V100 GPUs. 

\textbf{Inference.}
Consistent with stableSR, we implement DDPM sampling with 200 timesteps. However, fewer steps can yield comparable results, as discussed in Sec.\ref{subsec_4.3}. We employ evaluation metrics including LPIPS \citep{zhang2018unreasonable}, FID \citep{heusel2017gans},  MUSIQ \citep{ke2021musiq} and NIQE \citep{mittal2012making}. PSNR and SSIM scores are also reported on the luminance channel in the YCbCr color space.

\subsection{Benchmark Results of 4$\times$ Blind Image SR.}
\label{subsec_4.1}

\begin{table}
\begin{center}
\caption{Quantitative Results on 4$\times$ SR benchmarks. The 1st and the 2nd best performances are highlighted in red and blue, respectively. \textsuperscript{$\dagger$} means reproducing with the official model. 
}
\label{tab:4xSR}
\label{tab:main_table}
\setlength{\tabcolsep}{5pt}
\scalebox{0.7}{
\begin{tabular}{l|l|c|c|c|c|c|c|c|c|c}
\toprule[0.4mm]
Datasets & Metrics & RealSR 
 &  BSRGAN  & Real-ESRGAN+  &  DASR &  FeMaSR  &  LDM   & StableSR & StableSR\textsuperscript{$\dagger$} & OURS \\ \toprule[0.4mm]
    \multirow{7}*{\begin{tabular}{l}DIV2K \\ Valid\\
                \end{tabular}}  

 & PSNR$ \uparrow$  & 22.36 & \textcolor{blue}{22.71} & \textcolor{red}{22.93} & 22.70 & 21.73 & 21.86 & 23.26 & 21.81 & 22.11 \\ 
& SSIM$ \uparrow$ & 0.5559 & 0.5911 & \textcolor{red}{0.6144} & \textcolor{blue}{0.5988} & 0.5692 & 0.5554  & 0.5726  & 0.5534 & 0.5775 \\
 & LPIPS$ \downarrow$   & 0.6191 & 0.3546 & 0.3229 & 0.3545 & 0.3389 & 0.3264 & 0.3114  & \textcolor{blue}{0.3143} & \textcolor{red}{0.2821} \\
 & FID$ \downarrow$   & 71.85 & 49.0 & 40.49 & 51.79 & 41.26 & 27.36  & 24.44  & \textcolor{blue}{25.64} & \textcolor{red}{25.49} \\
 & MUSIQ$ \uparrow$  & 27.20 & 61.20 & 60.70 & 57.26 & 57.83 & 62.8906 & 65.92  & \textcolor{red}{66.78} & \textcolor{blue}{64.78} \\ 
& NIQE $ \downarrow$ & 7.71 & 4.84 & 4.88 & 4.95 & 4.95 & 5.64 
 & - & \textcolor{blue}{4.81} & \textcolor{red}{4.72} \\ 
\midrule
\multirow{6}*{\begin{tabular}{l}RealSR\\
                \end{tabular}} 
 & PSNR$ \uparrow$   & \textcolor{red}{27.77} & 26.55 & 25.82 & \textcolor{blue}{27.06} & 25.42 & 25.46  & 24.65 & 25.19 & 24.68 \\
 & SSIM$ \uparrow$ & \textcolor{blue}{0.7760} & 0.7742 & 0.7700 & \textcolor{red}{0.7833} & 0.7458 & 0.7145  & 0.7080 & 0.7227 & 0.7352 \\
 & LPIPS$\downarrow$   & 0.3805 & \textcolor{red}{0.2623} & \textcolor{blue}{0.2670} & 0.2923 & 0.2855 & 0.3159  & 0.3002 & 0.2974 & 0.2719 \\
 & FID$ \downarrow$  & 88.26 & 97.09 & 93.74 & 92.82 & 95.32 & 83.98  & - & \textcolor{red}{80.25} & \textcolor{blue}{83.59} \\
 & MUSIQ$ \uparrow$   & 29.88 & \textcolor{blue}{62.63} & 61.54 & 45.89 & 58.92 & 58.90  & 65.88 & \textcolor{red}{63.44} & 57.10 \\
 & NIQE $ \downarrow$ & 8.14 & \textcolor{red}{5.87} & 6.15 & 6.48 & \textcolor{blue}{5.90} & 6.78 & - & 6.35 & 5.96 \\ 
\midrule
\multirow{6}*{\begin{tabular}{l}DRealSR\\
                \end{tabular}} 
 & PSNR$ \uparrow$   & \textcolor{red}{31.73} & 29.98 & 29.69 & \textcolor{blue}{31.14} & 27.54 & 27.88 & 28.03 & 29.00 & 29.35 \\
 & SSIM$ \uparrow$ & \textcolor{red}{0.8563} & 0.8240 & 0.8277 & \textcolor{blue}{0.8493} & 0.7638 & 0.7448 & 0.7536 & 0.7658 & 0.7946 \\
 & LPIPS$\downarrow$  & 0.3634 & \textcolor{blue}{0.2757} & \textcolor{red}{0.2663} & 0.2873 & 0.3273 & 0.3379 &  0.3284 & 0.3353 & 0.3017 \\
 & MUSIQ $ \uparrow$ & 23.91 & \textcolor{blue}{55.12} & 52.21 & 40.50 & 52.56 & 53.72 & 58.51 & \textcolor{red}{56.72} & 42.32 \\
& NIQE $ \downarrow$ & 9.47 & \textcolor{blue}{6.70} & 7.04 & 7.84 & \textcolor{red}{6.22} & 7.37 & - & 6.89 & 6.89 \\ 
\bottomrule[0.4mm]
\end{tabular}
}
\end{center}
\end{table}

\begin{figure}
    \centering
    \includegraphics[width=1\linewidth]{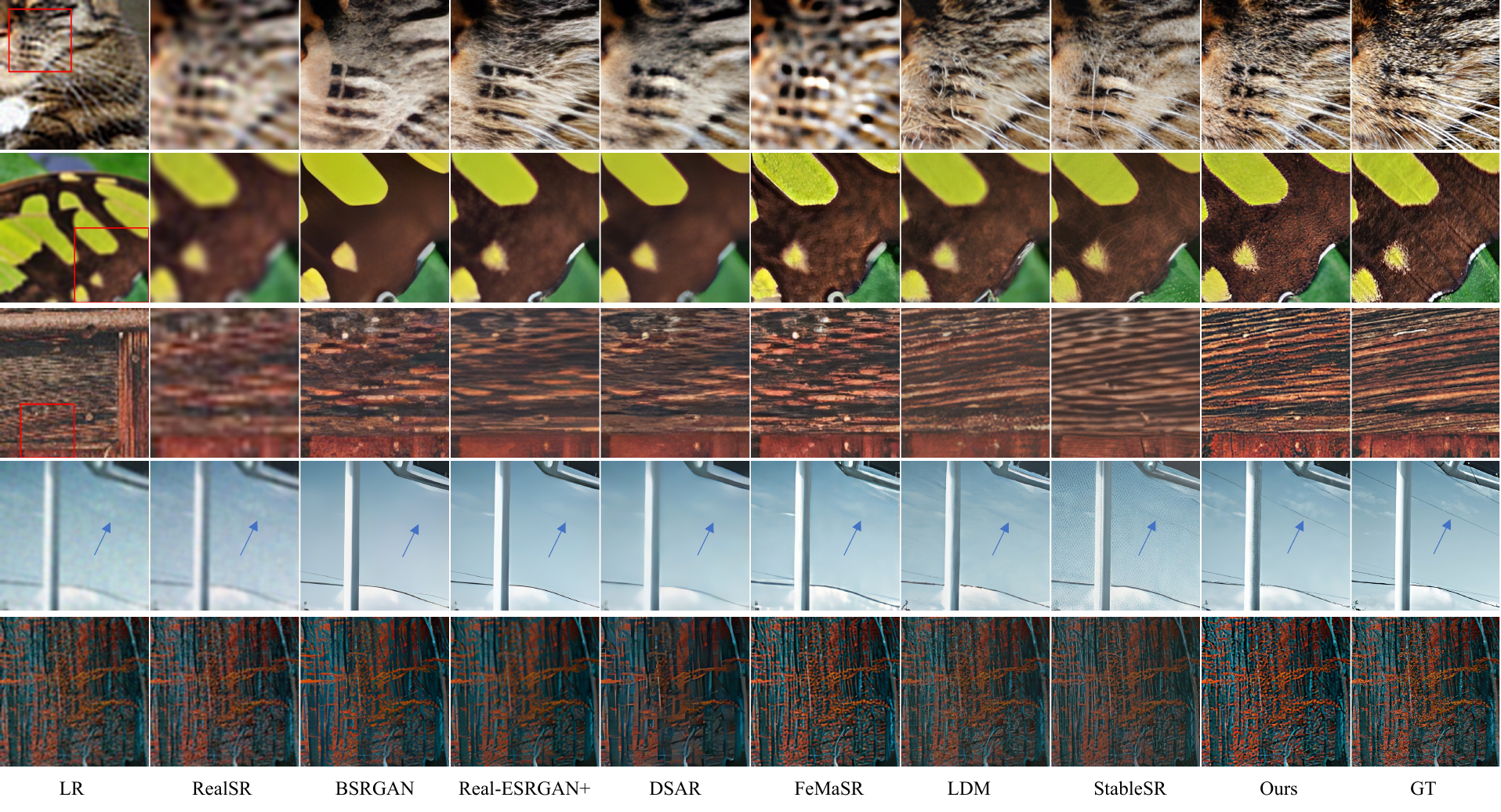}
    \caption{Qualitative comparisons on 4$\times$ SR (128 → 512). Our method is capable of achieving better detail consistency and generating realistic texture details. (Zoom in for the best view)}
    \label{fig:main_4xSR}
\end{figure}

We first evaluate our method on blind super-resolution. For synthetic data, we follow the degradation pipeline of Real-ESRGAN\citep{Wang2021RealESRGANTR} and generate $3k$ LR-HR pairs from DIV2K validation set. We compare our method quantitatively with GAN-based methods such as RealSR\citep{ji2020real}, BSRGAN\citep{zhang2021blind}, Real-ESRGAN+\citep{Wang2021RealESRGANTR}, DASR\citep{liang2022efficient}, FeMaSR\citep{chen2022real} and diffusion-based methods like LDM\citep{Rombach2021HighResolutionIS} and StableSR. The quantitative results are shown in Tab.~\ref{tab:4xSR}. Note that due to differences in making test sets, we reproduce StableSR using its official model and code. We can see that our approach outperforms state-of-the-art SR methods on perceptual metrics (including LPIPS, FID, and NIQE) and gets the best PSNR and SSIM among diffusion-based methods. Specifically, synthetic benchmark DIV2K Valid, our method achieves a 0.2821 LPIPS score, which is 10.24\% lower than StableSR and at least 12.64\% lower than other GAN-based methods. Besides, our method achieves the lowest LPIPS score among diffusion-based methods on the two real-world benchmarks \citep{cai2019toward, wei2020component}, which clearly demonstrates the superiority of our approach. Note that although GAN-based methods like BSRGAN and Real-ESRGAN+ achieve good MUSIQ and NIQE scores, but fail to restore faithful details, such as textures and small objects, and generate blurry results as shown in Fig.~\ref{fig:main_4xSR}. Compared with the diffusion-based methods, our method also produces more visually promising results by reserving more high-frequency information and achieving better detail consistency. 

\subsection{Benchmark Results of 8$\times$ Non-Blind Image SR.}
\label{subsec_4.2}

\begin{figure}
    \centering
    \includegraphics[width=1\linewidth]{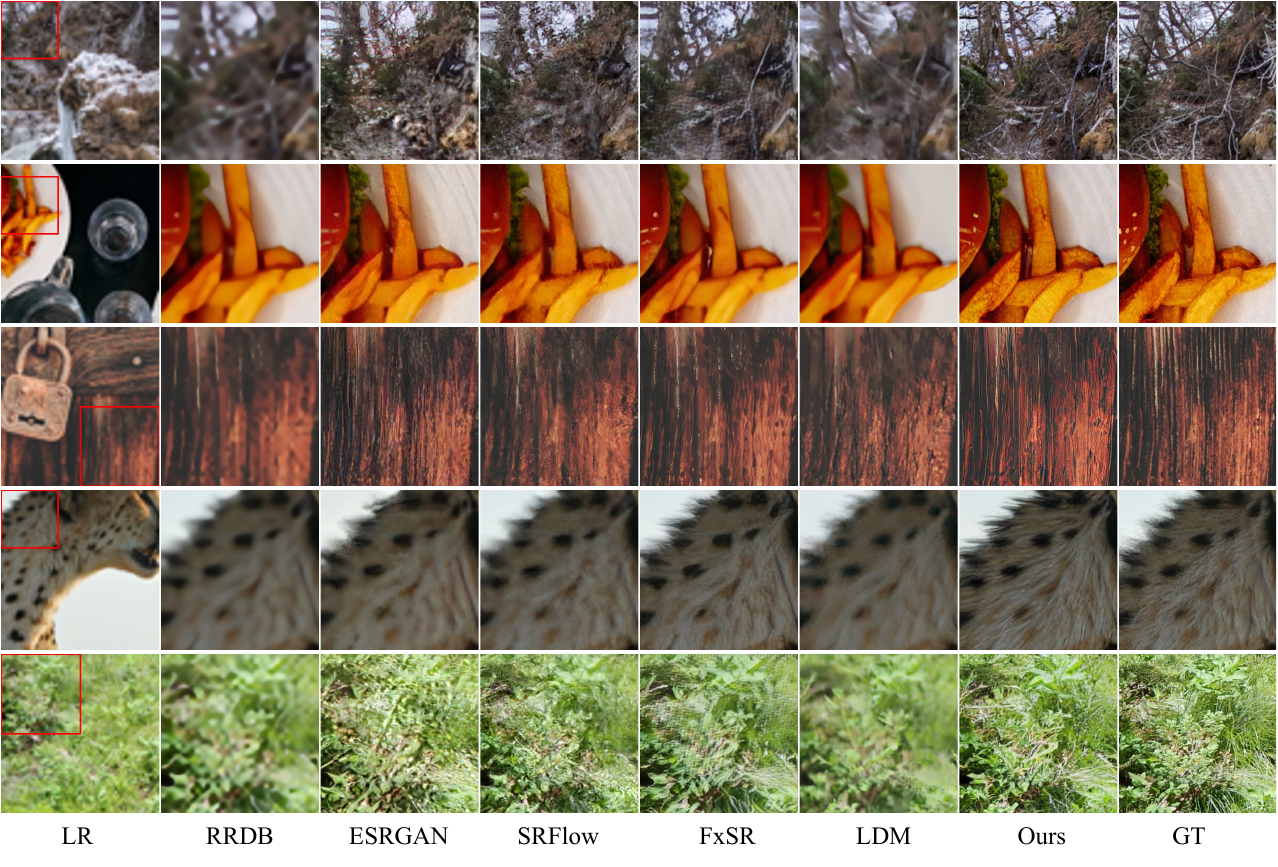}
    \caption{Qualitative comparisons on 8$\times$ SR (64 → 512). Our method is capable of generating sharp images with high-fidelity texture details. (Zoom in for the best view)}
    \label{fig:8x_main}
\end{figure}

We further validate the effectiveness of our method on 8$\times$ SR. For the test set, we generate 660 LR-HR pairs from DIV2K validation set via bicubic interpolation with the scale factor 8. We compare to other state-of-the-art models that span from regression models having powerful architectures and/or generative formulations: RRDB\citep{wang2018esrgan}, ESRGAN\citep{wang2018esrgan}, SRFLOW\citep{lugmayr2020srflow}, FxSR-PD\citep{park2022flexible}, LDM\citep{Rombach2021HighResolutionIS}. We use pre-trained models provided by the authors while for the non-provided 8$\times$ SR model (RRDB and ESRGAN), we get unofficial released models from the github\footnote{https://github.com/andreas128/SRFlow} of SRFLow. All results are tested on the same dataset using the official inference code. As Tab.~\ref{tab:8xSR} shows, our approach significantly outperforms competing methods on LPIPS, FID, and MUSIQ, and achieves top-2 in terms of NIQE. The qualitative results in Fig.~\ref{fig:8x_main} agree with the conclusions of the numerical results. It can be seen that our method can generate sharp images with high fidelity more naturally, while others tend to distort the characters or produce artifacts. Besides, our method can also generate realistic texture details, whereas others produce over-smooth results. Benefiting from the capacity and scalability of SS-MoE, our method has much more clear results compared to LDM's blurry output.
\begin{table}[h!]
\begin{center}
\caption{Quantitative Results on synthetic 8$\times$ SR benchmarks.}
\label{tab:8xSR}
\setlength{\tabcolsep}{5pt}
\scalebox{0.8}
{
\begin{tabular}{c|c|c|c|c|c|c|c|c}
\toprule[0.4mm]
Datasets & Metrics & Bicubic &  RRDB  & ESRGAN  &  SRFlow &  FxSR-PD  &  LDM   &  OURS \\
\toprule[0.4mm]
\multirow{6}{*}{ DIV2K Valid} & PSNR$ \uparrow$ & \textcolor{blue}{25.37} & \textcolor{red}{27.18} & 24.13 & 24.88 & 25.24 & 23.81 & 24.45 \\ 
& SSIM$ \uparrow$ & \textcolor{blue}{0.6361} & \textcolor{red}{0.6995} & 0.6035 & 0.6008 & 0.6312 & 0.5875 & 0.6142 \\
& LPIPS$ \downarrow$ & 0.6055 & 0.4332 & 0.2767 & 0.2706 & \textcolor{blue}{0.2425} & 0.3087 & \textcolor{red}{0.2321} \\
& FID$ \downarrow$  & 118.63 & 92.4 & 58.64 & 59.52 & \textcolor{blue}{55.0} & 63.07 & \textcolor{red}{44.49}  \\
& MUSIQ$ \uparrow$  & 20.80  & 46.43 & 55.84 & 55.20 & 62.10 & \textcolor{blue}{63.82} & \textcolor{red}{64.17} \\
& NIQE $ \downarrow$ & 11.31& 8.65 & \textcolor{red}{3.97} & 4.60 & 5.24 & 5.75 & \textcolor{blue}{4.34} \\ 
\bottomrule[0.4mm]
\end{tabular}
}
\end{center}
\end{table}

\subsection{Ablation Studies and Computation Cost Analysis}
\label{subsec_4.3}
\textbf{Ablation on SS-MoE. }We investigate the significance of our proposed Sampling-Space Mixture of Experts on both 4$\times$ and 8 $\times$ SR. Here, we use the original VAE to decode latents generated by different models. As shown in Tab.~\ref{tab:ablation moe}, the removals of Sampling MoE and Space MoE both lead to a noticeable performance drop in almost all evaluated metrics for different tasks, demonstrating both modules contribute to our approach's powerful generative ability and fidelity. 
Furthermore, we evaluate the models' performance under different sampling steps. As depicted in Tab.~\ref{tab:ablation_sample_step}, for each sampling step Sampling-MoE can generate high-resolution images with better perceptual quality, showing its stronger denoise ability  by modeling noises of different levels using multiple experts. We also notice that Sampling-MoE can achieve better FID and LPIPS with fewer steps. For example, the two metrics in both 4$\times$ and 8 $\times$ SR with $T=50$ outperform Space-MoE model with $T=200$, decreasing 75\% sampling steps and resulting in more efficient diffusion-based SR.

\begin{table}[t]
    \caption{The effectiveness of SS-MoE evaluated on both 4$\times$ and 8$\times$SR.}
    \label{tab:ablation moe}
    \scriptsize
    \centering
    \begin{adjustbox}{width=0.9\columnwidth}
    \begin{tabular}{p{1.9cm}|p{1.0cm}|p{1.3cm}|p{1.3cm}|p{1.0cm}|p{1.0cm}}
\toprule[0.4mm]  Model & PSNR$ \uparrow$  & SSIM $\uparrow$  & LPIPS$ \downarrow$     &  FID$ \downarrow$  & MUSIQ$ \uparrow$    \\
\toprule[0.4mm]  w/o Sampling MoE  &22.09/24.94 &0.5649/0.6176 & 0.3201/0.2497 & 26.21/46.34 &63.14/63.32 \\ 
w/o Space MoE & 22.15/24.91 & 0.5680/0.6175& 0.3134/0.2426 & 26.19/43.93 & 63.77/63.98 \\
w SS-MoE  & 22.25/25.16 &  0.5725/0.6273 & 0.3031/0.2267 & 23.61/41.37 & 64.06/63.61 \\
\bottomrule[0.4mm] 
\end{tabular}
\end{adjustbox}

\end{table}

\begin{table}[h!]
\caption{Comparison SS-MOE and Space-Moe on DIV2K with different sampling steps.}
\label{tab:ablation_sample_step}
    \scriptsize
    \centering
    \begin{tabular}{c|l|l|l|l|l|l|l}
\toprule[0.4mm]  Task& Model & Sampling Step& PSNR$ \uparrow$ & SSIM $\uparrow$ & LPIPS $\downarrow$&  FID$ \downarrow$ & MUSIQ$ \uparrow$\\
\toprule[0.4mm]   \multirow{10}{*}{4$\times$SR}&\multirow{4}{*}{Space-Moe}& T=200& 22.09& 0.5649& 0.3201& 26.21& 63.14\\ 
  && T=100& 22.26& 0.5730& 0.3188& 26.25& 63.15\\
  && T=50& 22.54& 0.5852& 0.3246& 28.42& 61.83\\
  && T=20& 23.05& 0.6051& 0.3547& 36.08& 57.39\\
 \cline{2-8}  &\multirow{4}{*}{SS-MoE}& T=200& 22.24& 0.5727& 0.3031& 23.61&64.06\\ 
  && T=100& 22.45& 0.5824& 0.3074& 24.15& 63.37\\
  && T=50& 22.70& 0.5918& 0.3125& 25.87& 62.12\\
  && T=20& 23.26& 0.6141& 0.3461& 34.58& 57.56\\ 
 \hline
 \multirow{10}{*}{8$\times$SR}& \multirow{4}{*}{Space-Moe}& T=200& 24.94& 0.6176& 0.2497& 46.34& 63.32 \\
 & & T=100& 25.16& 0.6262& 0.2495& 45.99& 62.64\\
 & & T=50& 25.47& 0.6381& 0.2595& 46.81& 61.36\\
 & & T=20& 26.07& 0.6606& 0.2829& 51.91& 57.49\\
 \cline{2-8}& \multirow{4}{*}{SS-MoE}& T=200& 25.16& 0.6273& 0.2267& 41.37& 63.61\\
 & & T=100& 25.37& 0.6365& 0.2302& 41.59& 63.05\\
 & & T=50& 25.60& 0.6449& 0.2361& 42.68& 62.06\\
 & & T=20& 26.20& 0.6676& 0.2668& 48.8& 58.41\\
\bottomrule[0.4mm] 
\end{tabular}
\end{table}

\begin{table}[t]
    \caption{Ablation Study of FCD.}
    \label{tab: ablation_decoder}
    \small
    \centering
    \begin{tabular}{c|c|c|c|c|l|l}
\toprule[0.4mm]  Model  & PSNR$ \uparrow$&  SSIM$ \uparrow$&  LPIPS$ \downarrow$&  FID$ \downarrow$&MUSIQ$ \uparrow$ &NIQE$ \downarrow$\\
\toprule[0.4mm]  Baseline& 22.25& 0.5725& 0.3038&  23.90& 64.02 &5.6093\\ 
 + AFF-Net& 22.16& 0.5805& 0.2808& 24.52& 63.66 &4.4337\\
 + FFL loss& 22.01& 0.5675& 0.2892& 24.94& 63.64 &4.5789\\
 + UNet + FFL& 22.19& 0.5791& 0.2856& 25.09& 63.94 &4.3652\\
 + AFF + FFL& 22.23& 0.5814& 0.2815& 24.30& 64.03 &4.3675\\
\bottomrule[0.4mm]
\end{tabular}
\end{table}

\begin{wraptable}{r}{0.6\linewidth}
    \caption{Parameter and computation cost comparison on 4$\times$ SR using 200 timesteps.}
    \label{tab:computation_cost}
    \small
    \centering
    \begin{adjustbox}{width=0.5\columnwidth}
    \begin{tabular}{c|c|c|c}
    \toprule[0.4mm] 
    Method  & Param (M) & \makecell{FLOPs(T) per step} & Total FLOPs(T)\\
    \toprule[0.4mm] 
     LDM & 168.95 & 0.1608 & 33.43 \\
     Ours & 605.30 & 0.1658 & 35.47 \\
     StableSR & 1409.11 & 0.4162 & 86.27 \\
\bottomrule[0.4mm]
\end{tabular}
\end{adjustbox}
\end{wraptable}

\textbf{Ablation on FCD.} Then, we aim to illustrate the effectiveness of our proposed Frequency Compensated Decoder. The ablation experiments in Table.~\ref{tab: ablation_decoder} have the same 25k training steps and are evaluated on 4$\times$ DIV2K. We use the VQ Model in LDM as the baseline and add AFF-Net and FFL Loss progressively. As shown in Table.~\ref{tab: ablation_decoder}, frequency refinement introduced by AFF Net and FFL loss improves the perceptual quality of images, with 7.3\% improvement of LPIPS and 22.1\% improvement of NIQE in contrast to baseline. Compared with UNet+FFL, AFF+FFL achieves lower LPIPS and FID, indicating better realism and suggesting the effectiveness of frequency operation.

\textbf{Parmeter and Computational Cost Analysis}
We further evaluate our method against other diffusion-based SR methods, including LDM and StableSR on 4$\times$ SR in terms of the parameter number and FLOPs. The results are shown in Tab.~\ref{tab:computation_cost}. 
We calculate FLOPs for one denoising step and a single SR inference separately and timesteps are set to 200 when inference. Notice that our model's parameter number, 605.30, includes SS-MoE with four experts and a frequency-augmented decoder. Benefitting from SS-MoE, the parameter number of our method increases by 436.35M and the FLOPs only increase by 3.1\% and 6.1\% compared with LDM. 
As for StableSR, it utilizes the stable diffusion 2.1 architecture along with a half UNet, resulting in approximately 2.5 times FLOPs compared to our method, highlighting our lightweight nature.

\section{Conclusion}
Unlike existing pixel diffusion-based SR methods that require huge calculating resources, we have introduced a latent diffusion model for efficient SR. We propose Sampling-Space MoE to enlarge the diffusion model without necessitating a substantial increase in training and inference resources. Furthermore, to address the issue of information loss caused by the latent representation of the diffusion model, we propose a frequency-compensated decoder to refine the details of super-resolution images. Extensive experiments on both Blind and Non-Blind SR datasets have demonstrated the superiority of our proposed method. 

\textbf{Limitations.}
While our method has demonstrated promising results, the potential of diffusion-based methods has not been fully explored. We encourage further exploration in Latent Diffusion SR to achieve stronger generalizability in real-world SR. Increasing the size of model and using more degradation pipelines of data may help alleviate the problem. Our frequency-compensation decoder does not completely address the distortion caused by latent space compression. Expanding the latent feature channel might be a solution to further increase the reconstruction accuracy, but it will also result in a model that is more difficult to converge.


\newpage

\bibliography{plainnat}
\bibliographystyle{references}

\newpage

\appendix
\section{Appendix}
\subsection{Implement Details}
In this part, we illustrate the details of our method, including model architecure and training setting. To be specific, the framework contains two parts, denoise UNet and Frequency Compensated Decoder, which correspond to training stage1 and training stage2 repsectively. The denoising UNet follows the architecture of Latent Diffusion Model and the Frequency Compensated Decoder is based on VQModel. All hyperparameters are as shown Tab.~\ref{tab: hyperparameter1} and Tab.~\ref{tab: hyperparameter2}.

\begin{table}[t]
    \caption{Hyper-parameters and values in SS-MoE and Frequency Compensated Decoder.}
    \label{tab: hyperparameter1}
    \centering
    \begin{tabular}{cc}
    \toprule[0.4mm] 
    Configs/Hyper-parameters & Values \\
    \toprule[0.4mm] 
    \textbf{SS-MoE} & \\
     $f$ & 4 \\
     $z$-shape & 35.47 \\
     Channels & 160  \\
     Channel multiplier & [1, 2, 4, 4]\\
     Attention resolutions & [16, 8] \\
     Head channels & 32\\
     Architectures of Space MoE & FFN \\
     Activations in Space MoE & GELU \\
     Number of Sampling MoEs & 4 \\
     Time stage split & [(1000, 750], (750, 500], (500, 250], (250, 0]] \\ 
     \\
     \textbf{Frequency Compensated Decoder} & \\
     Embed dim & 4 \\
     Number of embed & 8192 \\
     Double z & False \\
     Z channels & 3 \\
     Channels & 128 \\
     Channel multiplier & [1, 2, 4]\\
     Number of Residual blocks & 2 \\
     Attention resolutions & [ ] \\
     Dropout rate & 0.0 \\
     Number of feat fusion layers & 1 \\
     Number of AFF blocks & 1 \\
     \bottomrule[0.4mm]
\end{tabular}
\end{table}

\begin{table}[t]
    \caption{Hyper-parameters and values in two stages' training.}
    \label{tab: hyperparameter2}
    \centering
    \begin{tabular}{ccc}
    \toprule[0.4mm] 
    Configs/Hyper-parameters & Stage1 & Stage2 \\
    \toprule[0.4mm] 
    Loss & L2 & L1, LPIPS, GAN Loss, FFL Loss\\
    Training steps & 1e5 & 5e4 \\
    Learning rate  & 5e-5 & 1e-4 \\
    Batch size per GPU & 9 & 1 \\
    Accumulate grad batches & 2 & 4 \\
    Number of GPU & 8 & 8 \\
    GPU-type & V100-32GB & V100-32GB\\
    \bottomrule[0.4mm]
\end{tabular}
\end{table}

\subsection{More Qualitative Comparisons}
More qualitative Comparisons among diffusion-based SR methods are as follows.

\begin{figure}
    \centering
    \includegraphics[width=1\linewidth]{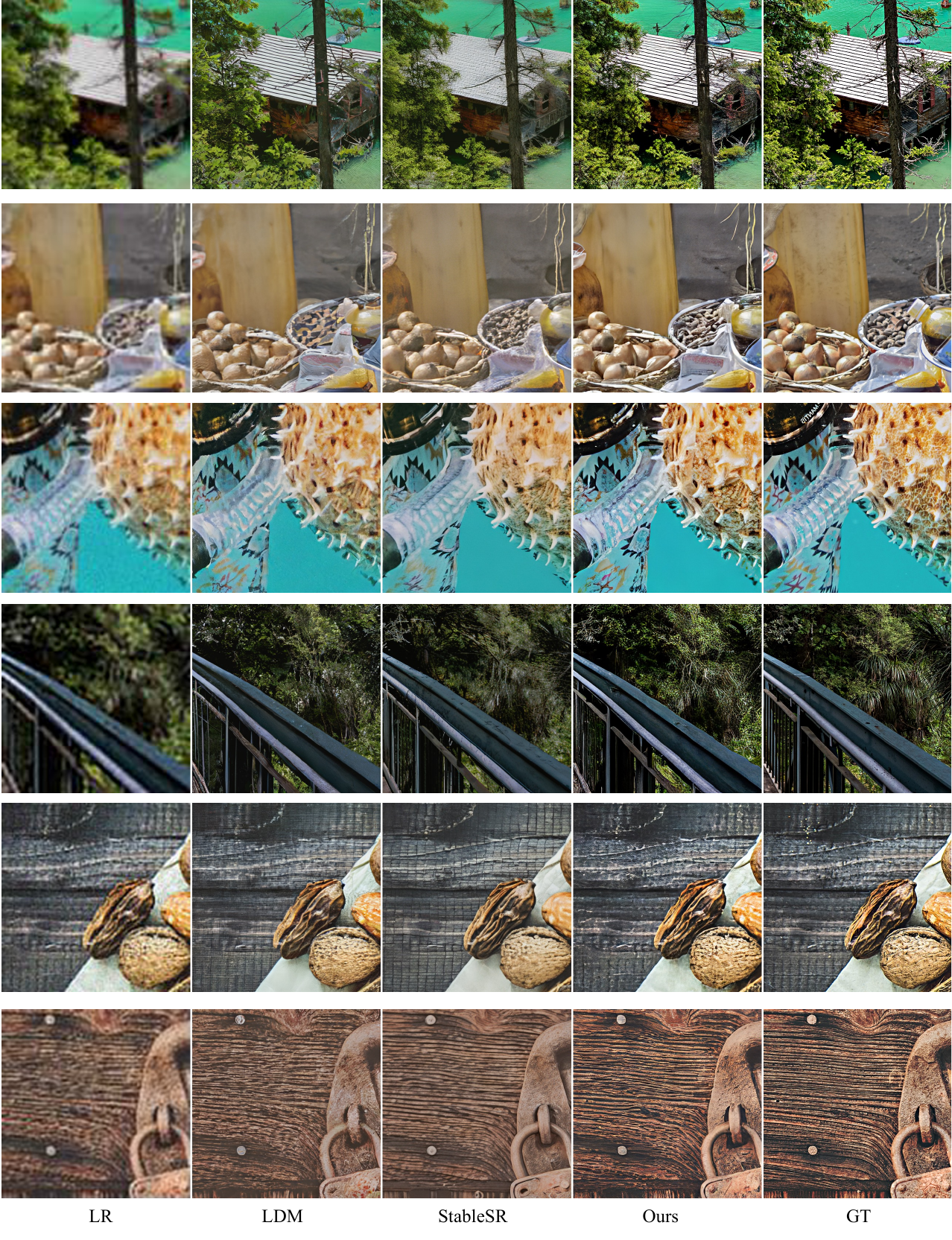}
    \caption{Qualitative comparisons on 4$\times$ SR (128 → 512). (Zoom in for the best view)}
    \label{fig:4x_add_main}
\end{figure}

\begin{figure}
    \centering
    \includegraphics[width=0.85\linewidth]{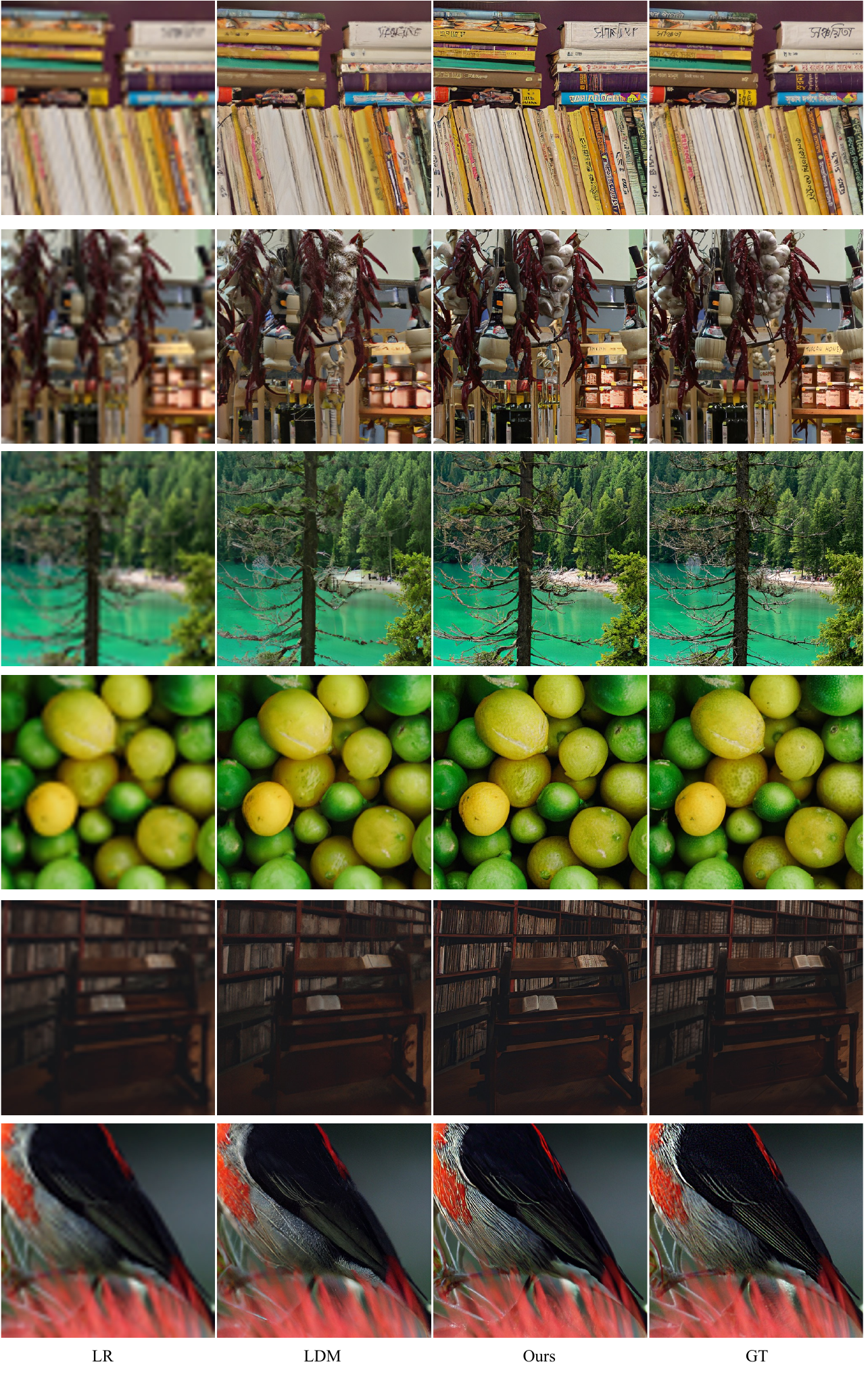}
    \caption{Qualitative comparisons on 8$\times$ SR (64 → 512). (Zoom in for the best view)}
    \label{fig:4x_add_main}
\end{figure}

\end{document}